\documentclass[12pt]{l4dc2022}
\let\subtable\relax
\newtheorem{problem}{Problem}
\usepackage{algorithm}
\usepackage{algpseudocode}
\DeclareMathOperator*{\argmax}{argmax}
\usepackage{wrapfig}
\usepackage{caption}

\title[Time-Incremental Learning]{Time-Incremental Learning from Data Using Temporal Logics}
\usepackage{times}

\author{%
 \Name{Erfan Aasi}$^{1}$ \Email{eaasi@bu.edu}\\
 \Name{Mingyu Cai}$^{2}$ \Email{mic221@lehigh.edu}\\
 \Name{Cristian Ioan Vasile}$^{2}$ \Email{crv519@lehigh.edu}\\
 \Name{Calin Belta}$^{1}$ \Email{cbelta@bu.edu}\\
 \addr $1$: Mechanical Engineering Department, Boston University, Boston, MA 02215, USA \\
 \addr $2$: Mechanical Engineering and Mechanics Department, Lehigh University, Bethlehem, PA 18015, USA
}

\begin{document}

\maketitle

\begin{abstract}%
Real-time and human-interpretable decision-making in cyber-physical systems is a significant but challenging task,
which usually requires predictions of possible future events from limited data. In this paper, we introduce a time-incremental learning framework: given a dataset of labeled signal traces with a common time horizon, we propose a method to predict the label of a signal that is received incrementally over time, referred to as prefix signal. Prefix signals are the signals that are being observed as they are generated, and their time length is shorter than the common horizon of signals. We present a novel decision-tree based approach to generate a finite number of Signal Temporal Logic (STL) specifications from the given dataset, and construct a predictor based on them. Each STL specification, as a binary classifier of time-series data, captures the temporal properties of the dataset over time. The predictor is constructed by assigning time-variant weights to the STL formulas. The weights are learned by using neural networks, with the goal of minimizing the misclassification rate for the prefix signals defined over the given dataset. The learned predictor is used to predict the label of a prefix signal, by computing the weighted sum of the robustness of the prefix signal with respect to each STL formula. The effectiveness and classification performance of our algorithm are evaluated on an urban-driving and a naval-surveillance case studies. 
\end{abstract}


\section{Introduction}\label{sec:introduction}

\begin{wrapfigure}{r}{0.50\columnwidth}
  \begin{center}
    \includegraphics[width=0.50\columnwidth]{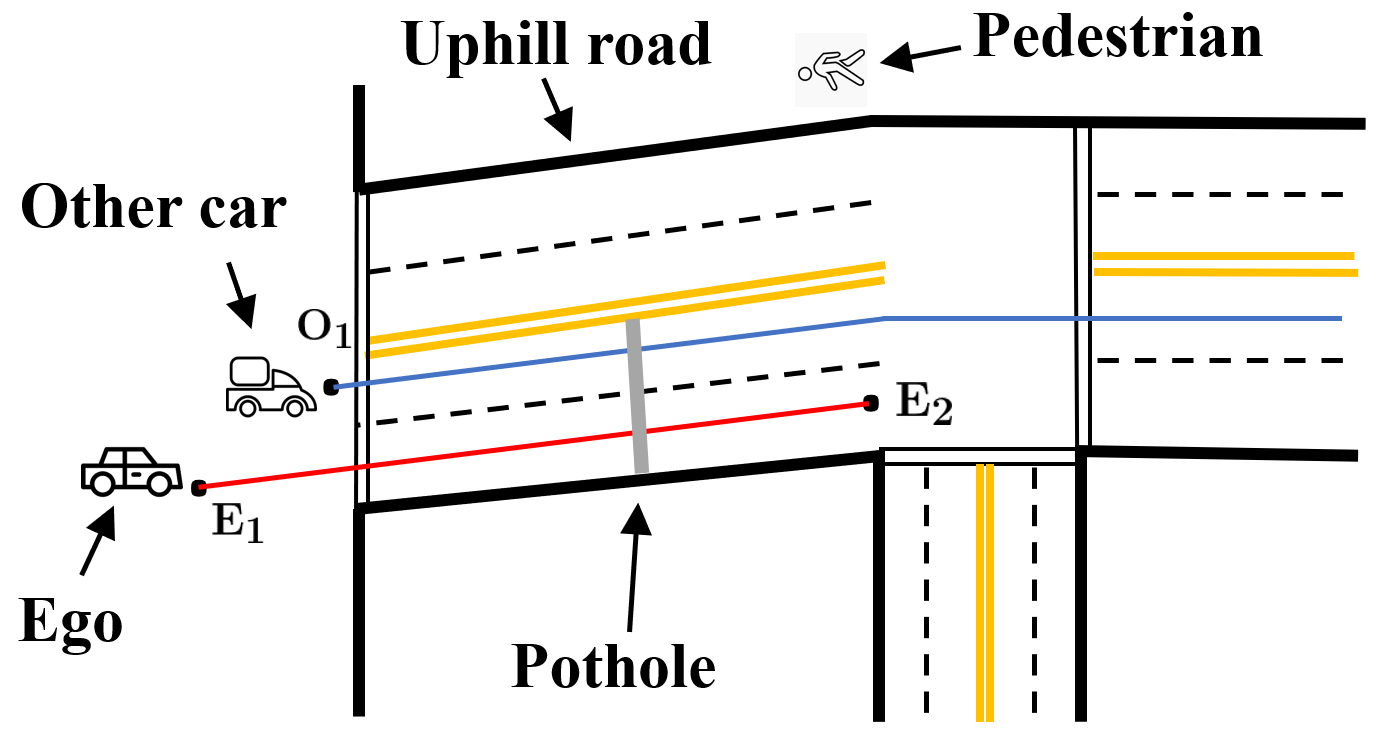}
  \end{center}
  \vspace{-5mm}
  \caption{Schematic of the urban-driving scenario. The vehicles are moving in a straight line in their lanes (red line for ego and blue line for the other car). The initial points of ego and other car are denoted by $E_1$ and $O_1$, and the goal point of ego is denoted by $E_2$, respectively.}
  \label{fig:car_scenario_schematic}
\end{wrapfigure}

Real-time decision-making for robotic tasks is a challenging problem, which usually requires prediction of possible outcomes in an incremental framework, based on available partial signals over time. The accuracy of such incremental predictions determines the efficiency of mitigating the occurrence of undesired behaviors. This is crucial in different fields, such as autonomous driving, e.g., in the case when a self-driving car does not have a clear vision over some objects in the environment and it has to take actions based on the performance of nearby cars, by observing their behavior over time. For example, consider the urban-driving scenario in Fig.~\ref{fig:car_scenario_schematic}. It consists of an autonomous vehicle, referred as {\em ego}, a pedestrian, and another car with a driver. The vehicles are headed toward an intersection with no traffic light, and there is an unmarked cross-walk at the end of the ramp-shape road, before the intersection. There are two possible types (labels) for the behavior of the other car: aggressive or safe. An aggressive driver keeps the same acceleration while moving in its lane, no matter whether the pedestrian crosses the street or not. A safe driver brakes slightly when it reaches the pothole, and if the pedestrian crosses the street, the driver applies full-brake to stop behind the intersection; otherwise, it keeps moving with the same acceleration. 

Predicting the behavior label of the human driver is valuable, especially in the case that ego does not have a clear line-of-sight to the pedestrian (because of the uphill shape of the road or presence of the other vehicle). In this situation, if the human driver is predicted as a safe driver, one possible strategy for ego is to follow its actions. In this paper, we focus on a time-incremental learning framework, where a dataset of the other car's behaviors (e.g., position and velocity) and their labels (safe or aggressive) that are recorded offline, is provided to ego. The goal is to develop a method allowing ego to predict whether the human driver is aggressive or safe, by observing its behavior on the fly.

For the same scenario, suppose the time horizon of the scenario is $T$. A dataset $S = \{s^i, \ell^i\}_{i=1}^{N}$, consisting of $N$ signals $s^i$ with time length $T$ and their corresponding labels $\ell^i$, is provided to ego. The signals are the recorded behavior of the other car over time and the labels represent the behavior type of the human driver. This is a two-class classification problem, and the goal is to develop a method allowing ego to classify the real-time behavior of the other car, represented by the prefix signal $s[0:t], 1 \leq t \leq T$.
In this paper, to provide interpretable specifications, we use Signal Temporal Logic (STL) \cite{maler2004monitoring} to specify the classifiers. 

We propose a novel framework to solve such a classification problem, which consists of three main parts and each part takes the given dataset as an input (see Fig.~\ref{fig:signal_analysis_and_diagram} (a)). The first part, named {\em "Signal Analysis"}, analyzes the signals and applies a heuristic method to find a finite number of timepoints along the horizon of signals, referred to as {\em decision times}. The decision times are the timepoints that are potentially informative  for separating the signals into two classes, and they are considered as candidate timepoints for generating classifiers. The next part, {\em "Classifier Learning"}, is responsible for generating classifiers at each decision time, using the given dataset. Here we use decision trees \cite{breiman1984classification}, \cite{ripley2007pattern} to learn the binary classifiers from data, and describe them by STL specifications. Each STL specification captures the temporal properties of the dataset over time. Lastly, by using Neural Networks (NNs) in the {\em "Classifier Evaluation"} part, we assign time-variant weights to the STL formulas, based on their classification performance for the prefix signals defined over the given dataset. The weighted conjunction of the STL formulas, interpreted as a weighted STL (wSTL) formula \cite{mehdipour2020specifying}, is considered as the output of "Classifier Evaluation", and the predictor is constructed based on that. The effectiveness and prediction power of our framework are evaluated on the above urban-driving and a naval surveillance case studies.

Note that a trivial solution to this problem is to apply an offline supervised learning method to the given dataset, learn an STL formula, and construct a monitor \cite{maler2004monitoring} to predict the label of prefix signals. The main limitation of this approach is that the output of a monitor is inconclusive for the prefix signals with time lengths shorter than the horizon of the learned STL formula, which makes it impractical for the time-incremental learning framework. Our approach addresses this limitation by providing predictions for all timepoints along the signal horizon.


\vspace{-2mm}

\subsection{Related Work}
Understanding the performance and detecting the desired behaviors of robotic systems from their execution traces have attracted a lot of attention recently. Most approaches are based on Machine Learning (ML) from data. 
Existing ML techniques usually construct a high-dimensional surface in the feature space, but they do not provide any insight on the meaning of the surfaces, which is particularly important for decision-making and prediction. This limitation has been recently addressed by integrating formal methods, and expressing classifiers as temporal
logic formulas \cite{clarke1986automatic} in
\cite{bartocci2014data}, \cite{mohammadinejad2020interpretable}, \cite{bombara2016decision}, \cite{xu2019information}, \cite{hoxha2018mining}, \cite{jha2019telex}, \cite{ketenci2019synthesis}, \cite{jin2015mining}, \cite{neider2018learning}, \cite{aasi2021classification}. Initial attempts for learning temporal logic properties from data focused on finding the optimal parameters for fixed formula structures \cite{bakhirkin2018efficient}, \cite{bartocci2015system} \cite{asarin2011parametric}, \cite{hoxha2018mining}, \cite{jin2015mining}. To learn both formula structure and its parameters, supervised classification methods are proposed, such as \cite{kong2016temporal} based on lattice search technique, and later in \cite{bombara2016decision} based on decision tree algorithm. There are other approaches in literature for learning temporal logic formulae, i.e., clustering \cite{vazquez2017logical}, \cite{bombara2017signal}, uncertainty-aware inference \cite{baharisangari2021uncertainty}, swarm STL inference \cite{yan2019swarm}, mining environment assumptions \cite{mohammadinejad2020mining}, and active learning \cite{linard2020active}. 

In \cite{aasi2021classification}, we proposed a boosted learning method, called \emph{Boosted Concise Decision Trees (BCDTs)}, for learning STL specifications, to improve on the classification performance and interpretability of existing works. The BCDT combines a set of shallow decision trees i.e. \emph{Concise Decision Trees (CDTs)}, which are empowered by a set of techniques to generate simpler formulae. The final output of the method is provided as a wSTL formula, which consists of the STL formulae from each CDT and their corresponding weights. Recent work \cite{yan2021neural} proposed a NN based method to learn the weights of a wSTL formula, from a given dataset. The neurons of their network correspond to subformulas and the output of neurons corresponds to the quantitative satisfaction of the formula.

\vspace{-2mm}
\section{Preliminaries}
\label{section:preliminaries}
Let $\mathbb{R}$, $\mathbb{R}_{\geq 0}$, $\mathbb{Z}_{\geq 0}$ represent the sets of real, non-negative real, and non-negative integer numbers, respectively.
Given $a,b\in \mathbb Z_{\geq0}$, we abuse the notation and use $[a,b] =  \{t\in\mathbb Z_{\geq 0}\ |\ a\leq t\leq b\}$.
A discrete-time signal $s$ with time horizon $T \in \mathbb{Z}_{\geq 0}$ is a function $s : [0, T] \to \mathbb{R}^n$
that maps each discrete time point $t \in [0, T]$
to an $n$-dimensional vector $s(t)$ of real values.
We denote the components of signal $s$ as $s_j, j \in [1, n]$, and the prefix of $s$ up to time point $t$ by $s[0{:}t]$. Let $\ell \in C = \{C_p, C_n\}$ denote the label of a signal $s$, where $C_p$ and $C_n$ are the labels for the positive and negative classes, respectively. We consider a labeled dataset with $N$ data samples as $S = \{s^i, \ell^i\}_{i=1}^{N}$, where $s^i$ is the $i^{th}$ signal and $\ell^{i}\in C$ is its corresponding label. A {\em prefix dataset $S[0:t_k]$ with horizon $t_k$}, is the dataset consisting of prefix signals with horizon $t_k$ and their labels, denoted by $\{s^i[0:t_k], \ell^i\}_{i=1}^{N}$. The cardinality of a set is shown by $|\cdot|$, an empty set is denoted by $\varnothing$, and a vector of zeros is denoted by $\emptyset$.

\textbf{Signal Temporal Logic (STL)}: 
STL was introduced in \cite{maler2004monitoring} to handle real-valued, dense-time signals. Informally, the STL specifications used in this paper are made of predicates defined over signal components in the form of $s_j(t) \sim \pi$, where $\pi \in \mathbb{R}$ is threshold and $\sim \in \{\geq, < \}$, which are connected using Boolean operators, such as $\neg$ ({\em negation}), $\wedge$ ({\em conjunction}), $\vee$ ({\em disjunction}), and temporal operators, such as $\mathbf{G}_{[a,b]}$ ({\em always}) and $\mathbf{F}_{[a,b]}$ ({\em eventually}). The semantics of STL are defined over signals. For example, formula $\phi_1 = \mathbf{G}_{[2,5]}s_1 < 4$ means that, for all times 2,3,4,5, component $s_1$ of a signal $s$ is less than 4, while formula $\phi_2 = \mathbf{F}_{[3,10]}s_2 > 4$ expresses that at some time between 3 and 10, $s_2(t)$ becomes larger than 4. 

STL has both qualitative (Boolean) and quantitative semantics. We denote Boolean satisfaction of a formula $\phi$ at time $t$ by $s(t) \models \phi$. For the quantitative semantics, the robustness degree \cite{donze2010robust}, \cite{fainekos2009robustness}, denoted by $\rho(s, \phi, t)$, captures the degree of satisfaction of a formula $\phi$ at time $t$ by a signal $s$. For simplicity of notation, we use $s \models \phi$ and $\rho(s, \phi)$ as short for $s(0) \models \phi$ and $\rho(s, \phi, 0)$, respectively. Boolean satisfaction $s \models \phi$ corresponds to non-negative robustness ($\rho(s, \phi) \geq 0$), while violation corresponds to negative robustness ($\rho(s, \phi) < 0$).  The minimum amount of time required to decide the satisfaction of a STL formula $\phi$ is called its horizon, and is denoted by $hrz(\phi)$. For example, the horizons of the two example formulas $\phi_1$ and $\phi_2$ given above are 5 and 10, respectively.

\textbf{Parametric STL (PSTL)}: PSTL \cite{asarin2011parametric} is an extension of STL, where the threshold $\pi$ in the predicates and the endpoints $a$ and $b$ of the time intervals in the temporal operators are parameters. A specific valuation of a PSTL formula $\psi$ under the parameter values $\theta \in \Theta$ is denoted by $\psi_\theta$, where $\Theta$ is the set of all possible valuations of the parameters.

\textbf{Weighted STL (wSTL)}: wSTL \cite{mehdipour2020specifying} is another extension of STL with the same qualitative semantics as STL, but its robustness degree is modulated by the weights associated with the Boolean and temporal operators. In this paper, we focus on a fragment of wSTL, with weights on conjunctions only. For example, for the wSTL formula $\phi_1 \wedge^{\alpha} \phi_2, \, \alpha=\left\{\alpha_1,\alpha_2\right\}\in \mathbb{R}^{2}_{\geq 0}$, for the case of $\rho(\phi_1, s), \rho(\phi_2, s) \leq 0$, its robustness is computed as $min(\alpha_1 \, . \, \rho(\phi_1, s), \, \alpha_2 \, . \, \rho(\phi_2, s))$, where the weights capture the importance of each formula in computing the robustness.


\section{Problem Formulation and Approach} \label{section:problemformulation} 
We first provide some definitions for the problem formulation.

\textbf{Predictor}:
The predictor $Pred_\Phi(s[0{:}t]) = \tilde{\ell}(t) \in C$ is a function that maps prefix signal $s[0:t]$ to a label $\tilde{\ell}(t) \in \{C_p, C_n\}$, which represents the satisfaction prediction of $s[0:t]$ at time $t$ with respect to the STL formula $\phi$. $Pred_\Phi(s[0:t]) = C_p$, if $s[0:t] \models \phi$; otherwise, $Pred_\Phi(s[0:t]) = C_n$.

\vspace{1mm}
\textbf{Timepoint MisClassification Rate (TMCR)}:
We define the TMCR at time step $t$, with respect to predictor $Pred_\Phi$, as below, where $\tilde{\ell}^i(t) = Pred_\Phi(s^i[0:t])$:
\begin{equation}
    TMCR(Pred_\Phi, t) = \sum_{i=1}^{N} \frac{|\{s^i[0:t] \mid (\tilde{\ell}^i(t) = C_p \, \wedge \, \ell^i = C_n ) \, \vee \, (\tilde{\ell}^i(t) = C_n \, \wedge \, \ell^i = C_p)\}|}{N} \nonumber
\end{equation}

\vspace{1mm}
\textbf{Incremental MisClassification Rate (IMCR)}:
The IMCR is defined as the vector of TMCR values over the time horizon of signals, denoted by:
\begin{equation}
    IMCR (Pred_\Phi) = [TMCR(Pred_\Phi, 0),\, TMCR(Pred_\Phi, 1), ...,\, TMCR(Pred_\Phi, T)] \nonumber
\end{equation}

\begin{problem} \label{probform:inc}
Given a labeled data set $S = \{(s^i, \ell^i)\}_{i = 1}^{N}$,
find an STL formula $\Phi$ and its corresponding predictor $Pred_\Phi$,
such that the $IMCR(Pred_\Phi)$ is minimized.
\end{problem}



\begin{figure}[htb]
\centering
\subfigure
{\centering
\includegraphics[width=0.40\columnwidth]{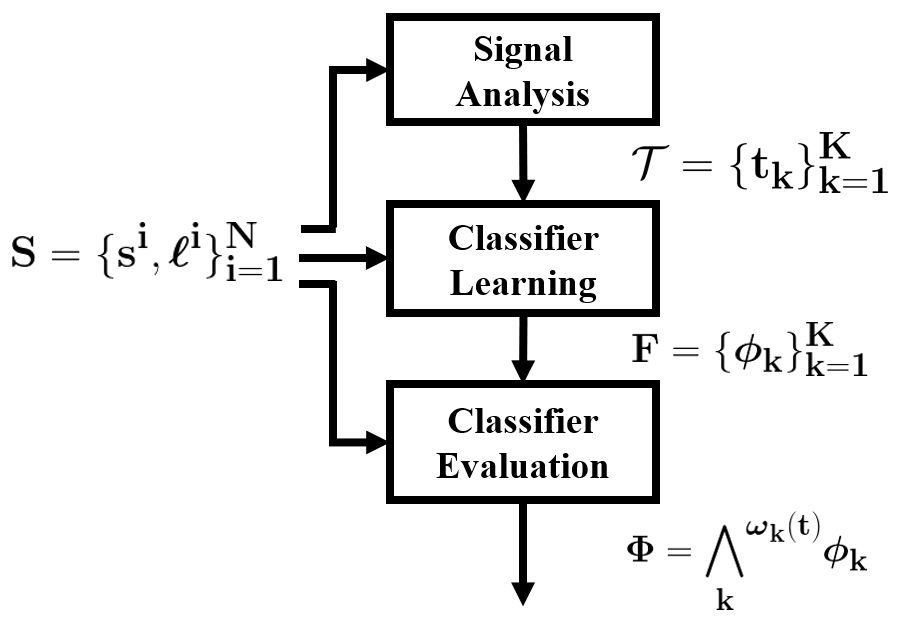}
\label{fig:learning_schematic}
}
\subfigure
{\centering
\label{fig:distance_naval}
\includegraphics[width=0.50\columnwidth]{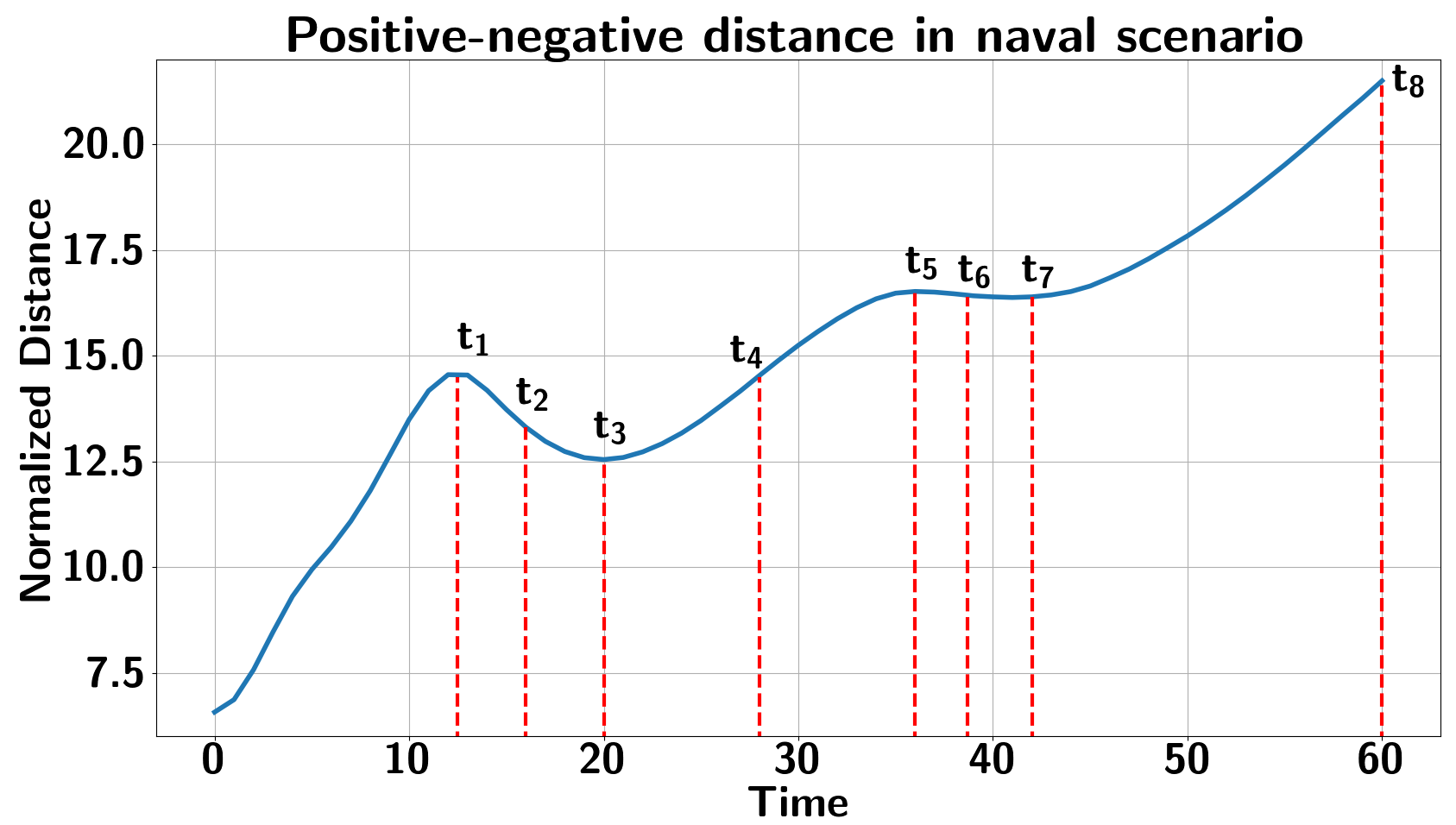}}
\caption{(a): Schematic of our framework, (b) Positive-negative distance in naval scenario. Based on the "Signal Analysis" part, the decision times are considered as $\mathcal{T} = \{t_1, t_2, ..., t_8\}$.}
\label{fig:signal_analysis_and_diagram}
\end{figure}

Our approach to Pb.~\ref{probform:inc} is illustrated in Fig.~\ref{fig:signal_analysis_and_diagram} (a). Our framework consists of three main components: (i) "Signal Analysis", described in Sec.~\ref{sec:decision_times}, applies a heuristic signal analysis method on the given dataset to find a finite number of potentially informative timepoints, referred as decision times and denoted by the set $\mathcal{T} = \{t_k\}_{k=1}^{K}$; 
(ii) "Classifier Learning", described in Sec.~\ref{sec:classifier_learning}, generates an STL formula for each decision time $t_k$ in $\mathcal{T}$ by using a decision tree method. The set of generated STL formulas is denoted by $F = \{\phi_k\}_{k=1}^{K}$; (iii) "Classifier Evaluation", explained in Sec.~\ref{sec:classifier_evaluation}, assigns a time-dependent weight distribution to the STL formulas in $F$, which captures the prediction performance of each STL formula over time. The output of this part is considered as the wSTL formula $\Phi = {\bigwedge_k}^{\omega_k(t)} \phi_k$ and a predictor $Pred_\Phi$ is constructed based on that.

We refer to our approach as a framework, because we provide a general framework consisting of a class of algorithms, where each algorithm can be replaced by other methods in literature, i.e., the heuristic method used in "Signal Analysis" part can be replaced by other signal analysis techniques in the literature, to find the decision times.

\vspace{-4mm}

\section{Signal Analysis}
\label{sec:decision_times}
Given the dataset $S$, the main role of the "Signal Analysis" part is to analyze the signals of the dataset over time and find a finite number of timepoints, referred to as decision times. The decision times, denoted by $t_k$, are the timepoints that are potentially informative for classifying the prefix dataset $S[0:t_k]$ into two classes, and they are considered as candidate timepoints for generating classifiers. Here we propose a heuristic method, based on the distance function among the signals, to find the decision times. 

A commonly used distance function for the case of multi-dimensional time-series data is the Euclidean distance. For two signals $s^1$ and $s^2$, the Euclidean distance \cite{bombara2017signal} is defined as $d^2 \, (s^1, s^2) = \sum_{j=1}^{n} \, \sum_{t = 0}^{T} \, (s^1_j(t) - s^2_j(t)) ^ 2$. Inspired by this, we define the {\em timepoint Euclidean distance} as $d^2 \, (s^1, s^2, t) = \sum_{j = 1}^{n} \, (s^1_j(t) - s^2_j(t)) ^ 2 , \, \forall t \in \{0, ..., T\}$. In the dataset $S$, the set of signals with positive labels are indexed by $h \in \{1, ..., N_p\}$ and with negative labels by $g \in \{1, ..., N_n\}$, respectively, such that $N_p + N_n = N$. We define the {\em positive-negative distance} in the dataset $S$ as:
\begin{equation} \label{eqn:positive_negative_distance}
    d^2_{pn} (t) = \sum_{j=1}^{n} \, \sum_{h = 1}^{N_p} \, \sum_{g = 1}^{N_n} \, (s^h_j(t) - s^g_j(t)) ^ 2  \quad , \quad \forall t \in \{0, ..., T\}
\end{equation}

We use~(\ref{eqn:positive_negative_distance}) as a metric to evaluate the separation between the positive and negative labeled signals from the dataset $S$ over time. As a simple, easy to compute heuristic method, we consider the decision times as the timepoints that the first- or second-order discrete derivatives of the function in~(\ref{eqn:positive_negative_distance}) are zero. Intuitively, these timepoints are the times that the positive and negative signals are locally at the furthest or the closest distance from each other (first-order derivatives), or the switching points for the evolution of the positive-negative distance over time (second-order derivatives). Also, we consider the horizon $T$ of signals as a decision time, to evaluate the whole traces of signals over time. The set of decision times is denoted by $\mathcal{T} = \{t_k\}_{k=1}^{K}$, where $\forall k \in \{1, ..., K\}: t_k \in \mathbb{Z}_{\geq 0}, \, 1 \leq t_k \leq T$. In Fig.~\ref{fig:signal_analysis_and_diagram} (b), the evolution of the positive-negative distance is depicted over time for the naval surveillance case study, described in Sec.~\ref{sec:naval_case_study}. Note that a trivial solution to Pb.~\ref{probform:inc} is to generate classifiers at every timepoint along the horizon of the signals, but obviously this is highly inefficient and computationally expensive. We compare the efficiency and the prediction accuracy of this trivial solution and our heuristic method in Sec.~\ref{sec:case_studies}.

\section{Classifier Learning} \label{sec:classifier_learning}

The next component, "Classifier Learning", takes as input the set of decision times $\mathcal{T}$ from the "Signal Analysis" part, in addition to the dataset $S$. "Classifier Learning" is responsible for generating classifiers at each decision time $t_k$ on the prefix dataset $S[0:t_k]$. To provide interpretable specifications for the classifiers and inspired by \cite{aasi2021classification, bombara2016decision}, we use the decision tree method $\mathcal{E}$ in Alg.~\ref{alg:dec_tree} to construct the classifiers. For each decision time $t_k \in \mathcal{T}$, Alg.~\ref{alg:dec_tree} is used to grow a decision tree $tree_k$, where each $tree_k$ is translated to a corresponding STL formula $\phi_k$. The structure of the method and its details are presented in \cite{bombara2016decision}, and here we explain it briefly.

The algorithm has three meta parameters: 1) PSTL primitives $\mathcal{P}$, where we use the first-order primitives described as $\mathcal{P}_1 = \{\mathbf{G}_{[t_0, t_1]} (s_j \sim \pi), \mathbf{F}_{[t_0, t_1]} (s_j \sim \pi)\}$, 2) impurity measure $\mathcal{J}$, where we use the extended misclassification gain impurity measure from \cite{bombara2016decision}, and 3) the stopping conditions $stop$, where we stop the growth of the trees when they reach a given depth. Alg.~\ref{alg:dec_tree} is recursive and takes as input (1) the prefix dataset $S[0:t_k] = \{s^i[0:t], \ell^i\}_{i=1}^{N}$, (2) the path formula to reach the current node $\phi^{path}$, and (3) the current depth level $h$.

At the beginning of the algorithm, the stopping conditions are checked (line~\ref{alg:line:stop}), and if they are satisfied, a single leaf that is assigned with label $c^* \in C$ is returned (lines~\ref{alg:line:leaf_label}-\ref{alg:line:leaf}), where $p(S[0:t_k], c; \phi^{path})$ captures the classification quality based on the impurity measure. If the stopping conditions are not satisfied, an optimal STL formula among all the possible valuations of the first-order primitives is found (line~\ref{alg:line:optimization}), denoted by $\phi^*$, and assigned to a new non-terminal node in the tree (line~\ref{alg:line:non_terminal}). Next, the prefix dataset $S[0:t_k]$ is partitioned according to the optimal formula, to the satisfying and violating prefix datasets $S_\top[0:t_k]$ and $S_\bot[0:t_k]$, respectively (line~\ref{alg:line:partition}), and the construction of the tree is followed on the left and right subtrees of the current node (lines~\ref{alg:line:tree_left}-\ref{alg:line:tree_right}). Note that the main difference of Alg.~\ref{alg:dec_tree} and the method in \cite{bombara2016decision}, is that the decision tree $tree_k$ constructed by Alg.~\ref{alg:dec_tree} is based on the prefix dataset $S[0:t_k]$ for each decision time $t_k \in \mathcal{T}$. Each decision tree $tree_k$ is translated to a corresponding STL formula $\phi_k$, using the translation method in \cite{bombara2016decision}. The output of the "Classifier Learning" is the set of STL formulas $F = \{\phi_k\}_{k=1}^{K}$, where $\forall \phi_k \in F: hrz(\phi_k) \leq t_k$.

\begin{algorithm}[htb]
\caption{Decision Tree Construction Method $\mathcal{E}$}
\begin{algorithmic}[1]
    \State \textbf{Meta-Parameters:} PSTL primitives $\mathcal{P}$, impurity measure $\mathcal{J}$, stopping conditions $stop$
    \State \textbf{Input:} prefix dataset $S[0:t_k]$, path formula $\phi^{path}$, current depth level $h$
    \State \textbf{Output:} sub-tree $tree_k$ 
    \State \textbf{if} $stop(\phi^{path}, h, S[0:t_k])$ \textbf{then} \label{alg:line:stop}
    \State \hskip1.5em $c^* = \argmax_{c \in C} \{p(S[0:t_k], c; \phi^{path})\}$ \label{alg:line:leaf_label}
    \State \hskip1.5em \textbf{return} $leaf(c^*)$ \label{alg:line:leaf}
    \State $\phi^* = \argmax_{\psi \in \mathcal{P}, \theta \in \Theta} \mathcal{J}(S[0:t_k], partition(S[0:t_k], \phi_\theta \wedge \phi^{path}))$ \label{alg:line:optimization}
    \State $tree_k \leftarrow non\_terminal(\phi^*)$ \label{alg:line:non_terminal}
    \State $S_{\top}[0:t_k], S_{\bot}[0:t_k] \leftarrow partition(S[0:t_k], \phi^{path} \wedge \phi^*)$ \label{alg:line:partition}
    \State $tree_k.left \leftarrow \mathcal{E}(S_{\top}[0:t_k], \phi^{path} \wedge \phi^*, h+1)$  \label{alg:line:tree_left}
    \State $tree_k.right \leftarrow \mathcal{E}(S_{\bot}[0:t_k], \phi^{path} \wedge \neg \phi^*, h+1)$ \label{alg:line:tree_right}
    \State \textbf{return} $tree_k$ $\label{alg:line:tree_output}$
\end{algorithmic}
\label{alg:dec_tree}
\end{algorithm}

\section{Classifier Evaluation} \label{sec:classifier_evaluation}
The final component of our framework, "Classifier Evaluation", takes as input the given dataset $S$ and the set of generated STL formulas $F = \{\phi_k\}_{k=1}^{K}$. "Classifier Evaluation" assigns a non-negative, time-variant weight distribution to the formulas, denoted by $\omega(t) = \{\omega_k(t)\}_{k=1}^{K}$, based on the classification performance of the formulas over time. In Alg.~\ref{alg:nn_weights}, we present a method to find the weights of the formulas $\omega(t)$ over time. With slight abuse of notation, we denote the column vector of the weights at time $t$ by $\omega(t)$, which is a $K \times 1$ array. In Alg.\ref{alg:nn_weights}, we desire to find the $K \times T$ dimensional matrix $\Omega = [\omega(0), \, \omega(1), ..., \, \omega(T)]$ that includes the vectors of the weights over time.

First, we introduce some notations: at each time step $t$, the subset of the formulas in $F$ that have a horizon less than or equal $t$ is denoted by $F_t^{\leq}$, and the rest of the formulas that have higher horizon are denoted by $F_t^>$. Note that $F_t^{\leq} \cap F_t^> = \varnothing$ and $F_t^{\leq} \cup F_t^> = F$. The main reason for this partitioning is that at each time step $t$ and for a prefix signal $s^i[0:t]$, the formulas in $F_t^{\leq}$ are able to predict a label for $s^i[0:t]$, based on their satisfaction or violation with respect to the prefix signal. However, the set of formulas in $F_t^>$ may not be conclusive enough to predict a label and they need more time instances of the prefix signal. It is clear that at $t=0, F_0^{\leq} = \varnothing$ and $F_0^> = F$, and at $t = T$, $F_T^{\leq} = F$ and $F_T^> = \varnothing$.

Alg.~\ref{alg:nn_weights} takes as input the dataset $S$ and the set of STL formulas $F$. The subsets $F_0^{\leq}$ and $F_0^>$ and the weight vector $\omega(0)$ are initialized by $\varnothing, F$, and the zero vector $\emptyset$, respectively (line~\ref{nn:line:initialize}). For each time step $t$ along the horizon of signals (line~\ref{nn:line:for_loop}), the subsets of formulas $F_t^{\leq}$ and $F_t^{>}$ are computed by the $partition\_formulas$ function (line~\ref{nn:line:partition_formulas}). This function compares the horizons of the formulas in $F$ with the current time step $t$, and partitions them into $F_t^{\leq}$ and $F_t^>$. If there is no update in the subset $F_t^{\leq}$ compared to the previous time step (line~\ref{nn:line:same_formulas}), the same weight vector from the previous time step is used for the current time (line~\ref{nn:line:same_omega}). If the set $F_t^{\leq}$ is updated (line~\ref{nn:line:otherwise}), first, we construct the prefix dataset $S[0:t]$ (line~\ref{nn:line:prefix_dataset}). Then, the robustness of the prefix signals in $S[0:t]$ are computed with respect to the formulas in $F_t^{\leq}$, by the $compute\_robustness$ function, and stored in the robustness matrix $R_t^{\leq}$ (line~\ref{nn:line:compute_robustness}). The dimensions of the $R_t^{\leq}$ are $N \times |F_t^{\leq}|$, where the $i^{\text{th}}$ row contains the robustness of prefix signal $s^i[0:t]$ with respect to the formulas in $F_t^{\leq}$. The robustness matrix $R_t^{\leq}$ and the labels of the signals $\{\ell^i\}_{i=1}^{N}$ are used to learn the weights of the formulas in $F_t^{\leq}$ at time $t$, denoted by the vector $\omega^{\leq}(t)$, and the weights of the formulas in $F_t^>$ are considered as zero, denoted by $\omega^>(t)$ (line~\ref{nn:line:learn_weights}). The function $learn\_weights$ constructs a Neural Network (NN) to learn the weights of the formulas in $F_t^{\leq}$. Using NNs to learn the weights of wSTL formulas has been explored previously in \cite{yan2021neural}. Inspired by \cite{cuturi2017soft}, the differentiable loss function of our designed NN measures the difference between the predicted label of the prefix signal $s^i[0:t]$, and its actual label $\ell^i$ from the dataset $S$. Finally, the weight vector $\omega(t)$ is transformed to a column vector (line~\ref{nn:line:all_weights}), and added to the weight matrix $\Omega$.


\begin{algorithm}[htb]
\caption{Learning the weights of the STL formulas}
\begin{algorithmic}[1]
    \State \textbf{Input:} dataset $S = \{s^i, \ell^i\}_{i=1}^{N}$, set of STL formulas $F = \{\phi_k\}_{k=1}^{K}$ 
    \State \textbf{Output:} matrix of the weights $\Omega$
    \State \textbf{Initialize:} $F_0^{\leq} \leftarrow \varnothing, \, F_0^> \leftarrow F, \, \omega(0) \leftarrow \emptyset$ \label{nn:line:initialize}
    \State \textbf{For} $t= 1,..., T:$  \label{nn:line:for_loop}
    \State \hskip1.5em $F_t^{\leq}, \, F_t^{>} \leftarrow partition\_formulas(F, t)$ \label{nn:line:partition_formulas}
    \State \hskip1.5em \textbf{if} $F_t^{\leq} = F_{t-1}^{\leq}$ \textbf{then} \label{nn:line:same_formulas}
    \State \hskip1.5em \hskip1.5em $\omega(t) = \omega(t-1)$ \label{nn:line:same_omega}
    \State \hskip1.5em \textbf{Otherwise} \label{nn:line:otherwise}
    \State \hskip1.5em \hskip1.5em $S[0:t] = \{s^i[0:t], \ell^i\}_{i=1}^{N}$ \label{nn:line:prefix_dataset}
    \State \hskip1.5em \hskip1.5em R$_t^{\leq} \leftarrow compute\_robustness(F_t^{\leq}, S[0:t])$ \label{nn:line:compute_robustness}
    \State \hskip1.5em \hskip1.5em $\omega^{\leq}(t) \leftarrow learn\_weights
    (R_t^{\leq}, \{\ell\}_{i=1}^{N}) \, , \quad \omega^{>}(t) \leftarrow \emptyset$ \label{nn:line:learn_weights} 
    \State \hskip1.5em \hskip1.5em $\omega(t) = [\omega^{\leq}(t), \omega^{>}(t)]^\top$ \label{nn:line:all_weights}
    \State \textbf{return} $\Omega = [\omega(0), \, \omega(1), ..., \, \omega(T)]$ $\label{nn:line:output}$
\end{algorithmic}
\label{alg:nn_weights}
\end{algorithm}

The output of the "Classifier Evaluation" is considered as the weighted conjunction of the STL formulas in $F$, denoted by the wSTL formula $\Phi = {\bigwedge_k}^{\omega_k(t)} \phi_k$. The final output of our framework is considered as $Pred_\Phi$, which predicts the label of a prefix signal $s^i[0:t]$ as:
\begin{equation}
    Pred_\Phi(s^i[0:t]) = \tilde{\ell}^i(t) = sign \, (\sum_{k=1}^{K} \omega_k(t) \, . \, \rho(s^i[0:t], \, \phi_k)).\label{eq:predictor}
\end{equation}
Note that our proposed predictor computes the robustness of the prefix signal as the weighted sum of the robustnesses of each STL formula in $\Phi$, which is different from monitoring the wSTL formula $\Phi$ and computing its robustness by the methods in \cite{mehdipour2020specifying,yan2021neural}. The time-dependent nature of the weights adjusts the predictor, based on the classification performance of the STL formulas and the time length of the prefix signals. In Sec.~\ref{sec:case_studies}, we emphasize on the importance of the weight distributions of the STL formulas, by showing the performance of the predictor, with and without considering the weight distributions.

\textbf{Remark:} Note that for a formula $\phi^k$ with $hrz(\phi_k)$, the robustness $\rho(s^i[0:t], \phi_k)$ of a prefix signal $s^i[0:t]$ is constant for all $t \geq hrz(\phi_k)$. Therefore, whenever there is an update in $F_t^{\leq}$ compared to $F_{t-1}^{\leq}$, the first $|F_{t-1}^{\leq}|$ columns of the $R_t^{\leq}$ are equal to $R_{t-1}^{\leq}$, and the robustness computations are done for the last $|F_t^{\leq}| - |F_{t-1}^{\leq}|$ columns of $R_t^{\leq}$. This technique improves the efficiency and computational time of our method.

\vspace{-2mm}
\section{Case Studies} \label{sec:case_studies}
We demonstrate the usefulness and classification performance of our approach with two case studies. The first one is the urban-driving scenario from Fig.~\ref{fig:car_scenario_schematic}, implemented in the simulator CARLA \cite{dosovitskiy2017carla}. The second is the naval surveillance scenario from \cite{kong2016temporal}. We compare our framework with two baseline methods. In the first one, referred by {\em all-times}, instead of choosing a finite number of decision times, we generate classifiers at all time points along the horizon $T$ of signals. We compare our framework with this baseline method, in the sense of efficiency and classification performance, and show the significance of choosing a finite number of decision times. In the second baseline, referred by {\em uniform-weights}, we consider an uniform distribution for the weights of formulas at all decision times. The main purpose of comparing our framework with uniform-weights baseline is to emphasize on the importance of considering time-variant weight distributions in the predictor function.

We use Particle Swarm Optimization (PSO) \cite{kennedy1995particle} to solve the optimization problems in Alg.~\ref{alg:dec_tree}. The parameters of the PSO and the NN used in Sec.~\ref{sec:classifier_evaluation}, are tuned empirically. In both scenarios, we evaluate our framework with maximum depth of 2 for the decision trees, and with 3-fold cross validation. The execution times are measured based on the system's clock. All computations are done in Python 2, on an Ubuntu 18.04 system with an Intel Core i7 @$3.70$ GHz processor and $16$ GB RAM.

\textbf{Remark:} In Alg.~\ref{alg:nn_weights}, in the early time steps $t$ when none of the formulas in $F$ have a horizon less than $t$, the set $F_t^{\leq}$ is empty and the weights of all formulas in $F$ are zero. In such time steps, for the evaluation of our framework, we assign a value of $0.5 (50\%)$ to $TMCR(Pred_{\Phi}, t)$. We make this convention because the predictor is inconclusive and we force it to predict a label based on i.i.d. fair coin tosses.
\vspace{-2mm}

\subsection{Urban-driving scenario} \label{sec:urban_driving}
Consider the urban-driving scenario from Sec.~\ref{sec:introduction} and depicted in Fig.~\ref{fig:car_scenario_schematic}. Ego and the other car are in different, adjacent lanes, moving in the same direction on an uphill road, by applying constant throttles. The throttle of ego is smaller than that of the other car, and the positions of the cars are initialized such that the other car is always ahead of ego. There is a pothole in the middle of the uphill road. We implement this scenario in the simulator CARLA. In our implementation, the cars move uphill in the $y-z$ plane of the coordinate frame, towards positive $y$ and $z$ directions, with no lateral movements in the $x$ direction. The simulation ends whenever ego gets closer than $2m$ to its goal point. We assume ego is able to estimate the relative position and velocity of the other car. The dataset of the scenario consists of 300 signals with 477 uniform time-samples per trace. 150 of the signals are for an aggressive driver and 150 are for a safe driver. The signals are 4-dimensional, which are the relative position and velocity of the other car in $y$ and $z$ axes. 

\begin{figure}[htb]
\centering
\subfigure[]
{\centering 
\label{fig:carla_imcr}
\includegraphics[width=0.49\columnwidth]{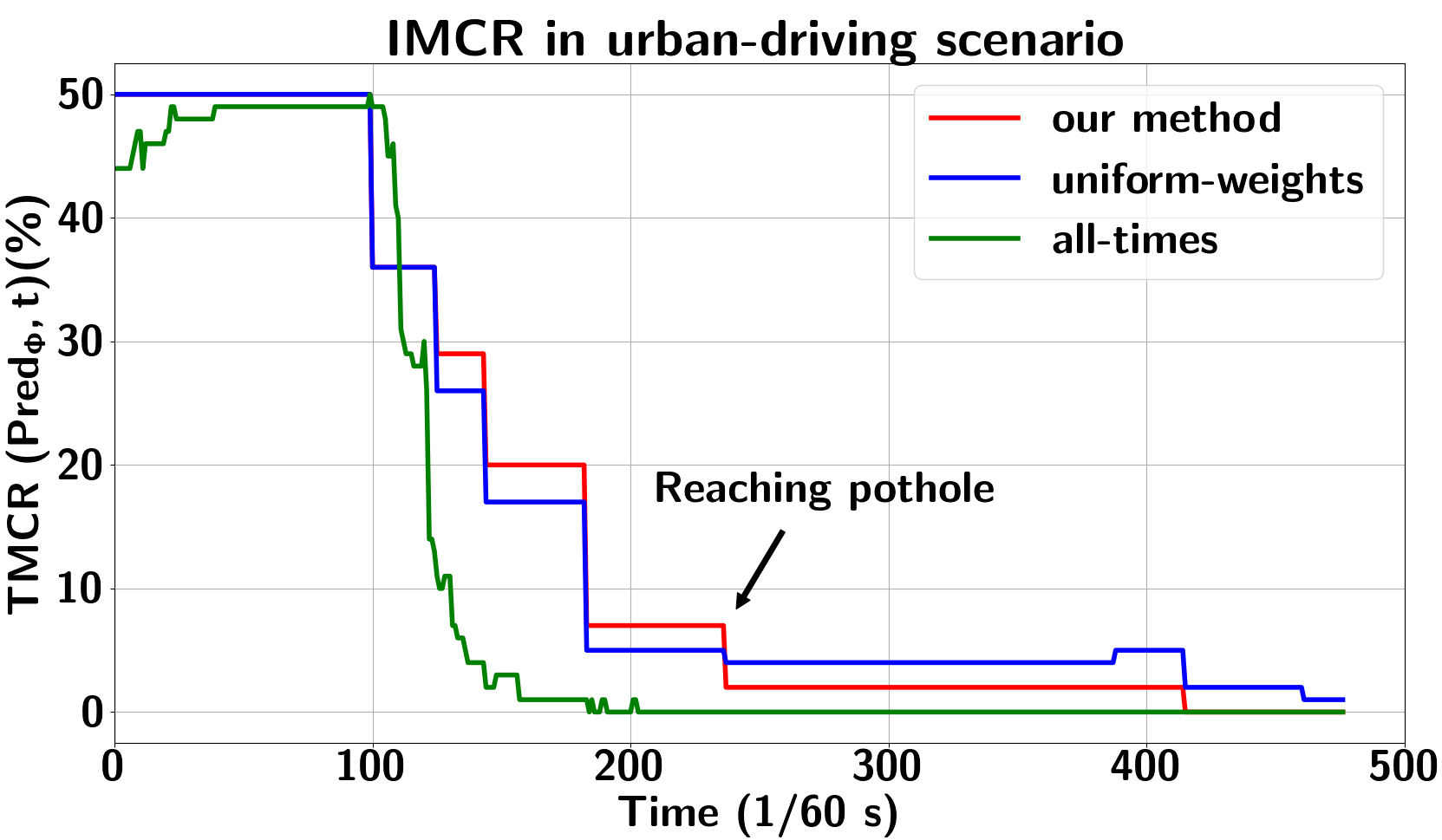}}
\subfigure[]
{\centering
\label{fig:naval_imcr}
\includegraphics[width=0.49\columnwidth]{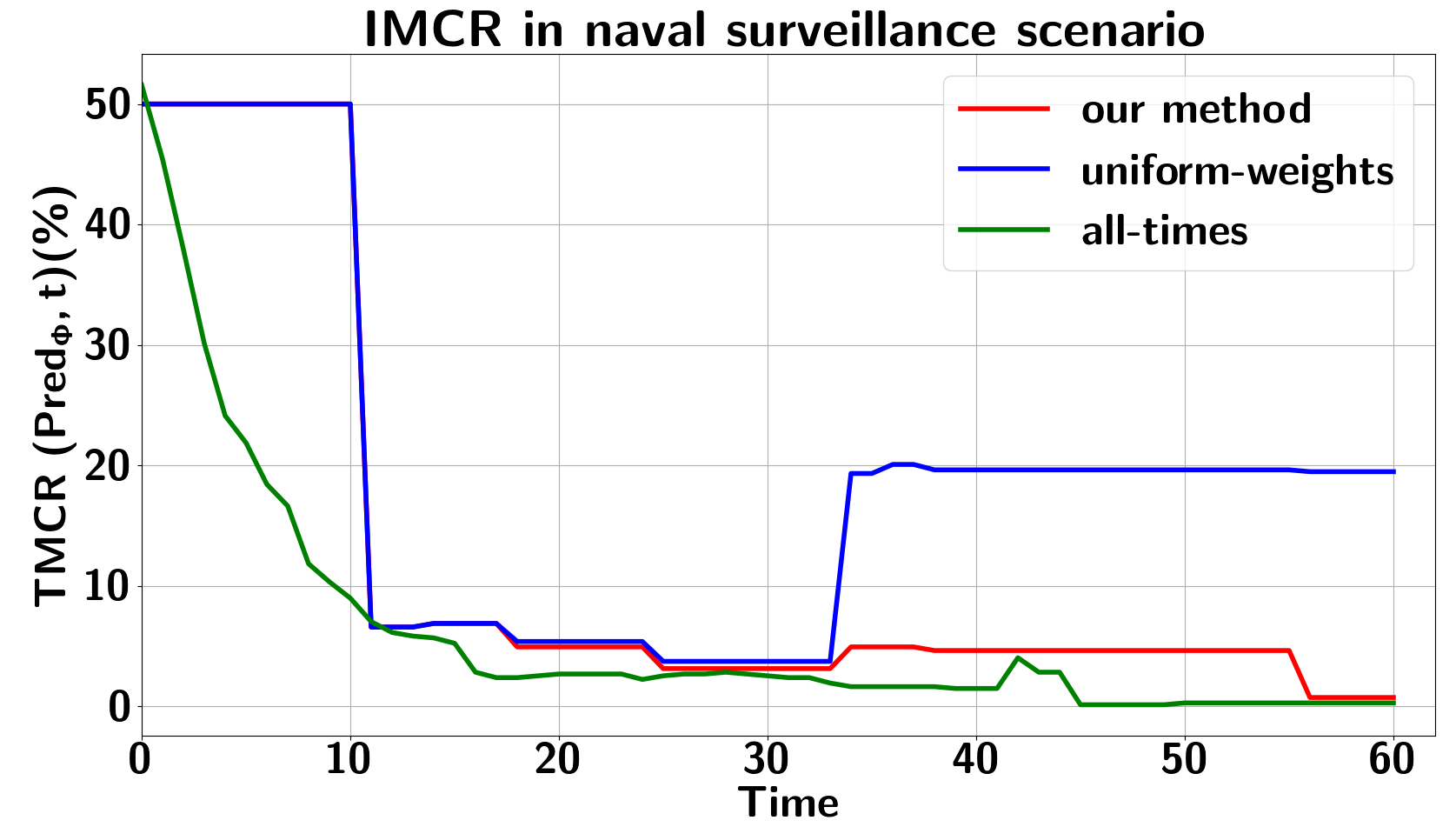}}
\caption{IMCR comparison of our framework with the baseline methods, in (a) urban-driving, and (b) naval surveillance scenarios.}
\label{fig:both_imcr}
\end{figure}

The "Signal Analysis" part of our framework finds 9 decision times for the dataset, and the wSTL formula from the "Classifier Evaluation" part is $\Phi = {\bigwedge_{k=1}^{9}} ^ {\omega_k(t)} \phi_k$. As an example formula in one the folds, the STL formula learned for the decision time $t_1 = 101$ is: $\phi_1 = (\phi_{11} \, \wedge \, \phi_{12}) \, \vee \, (\neg \, \phi_{11} \, \wedge \, \phi_{13})$, where $\phi_{11} = F_{[94, 100]} (v_y \leq 1.12)$, $\phi_{12} = G_{[77, 99]} (y \leq 11.03)$, and $\phi_{13} = G_{[85, 88]} (y \leq 10.59)$. For example, $\phi_{11}$ states that at some timepoint between 94 to 100, the relative velocity of the other car in y-axis gets less than or equal to 1.12 $m/s$. The IMCR comparison of our method with the baseline methods, by the same initialization of the parameters, is shown in Fig.~\ref{fig:both_imcr} (a). The runtime of our framework, the uniform-weights, and the all-times baselines are $646s$, $671s$, and $33158s$, respectively. Note that in this scenario, when the other car reaches the pothole, the behavior of safe and aggressive drivers are different. In Fig.~\ref{fig:both_imcr} (a), although the IMCR of our framework is close to the uniform-weights method before reaching the pothole, the IMCR of our approach shows better classification performance than the uniform-weights baseline after that. For the all-times baseline, its IMCR is generally better than our framework, over the horizon of signals, but its runtime is drastically larger. Moreover, the all-times baseline generates 477 formulas for this scenario and learns their corresponding weight distributions, which requires exponentially larger memory than our framework that only has 9 formulas.

\vspace{-5mm}

\subsection{Naval scenario} \label{sec:naval_case_study}
The naval surveillance problem was proposed in \cite{kong2016temporal}, based on the scenarios from \cite{kowalska2012maritime}. The goal of the scenario is to detect the anomalous vessel behaviors from their trajectories. Normal trajectories belong to the vessels that approach from the open sea and head directly toward the harbor. Anomalous trajectories belong to the vessels that either veer to the island and then head to the harbor, or they approach other vessels in the passage between the peninsula and the island and then veer back to the open sea. The dataset of the scenario consists of 2000 signals, with 1000 normal and 1000 anomalous trajectories. The signals are represented as 2-dimensional trajectories with planar coordinates ($x(t), y(t)$), and they have 61 timepoints. The labels indicate the type of the vessel's behavior (normal or anomalous).

The "Signal Analysis" component of our framework finds 8 decision times (see Fig.~\ref{fig:signal_analysis_and_diagram} (b)), and the final wSTL formula is $\Phi = {\bigwedge_{k=1}^{8}}^{\omega_k(t)} \phi_k$. As an example formula in one of the folds, the learned STL formula for the decision time $t_3 = 20$ is $\phi_3 = (\phi_{31} \, \wedge \, \phi_{32}) \, \vee \, (\neg \, \phi_{31} \, \wedge \, \phi_{33})$, where $\phi_{31} = G_{[11, 16]} (y > 23.33)$, $\phi_{32} = F_{[15, 18]} (y \leq 33.83)$, and $\phi_{33} = G_{[4, 13]} (x > 42.69)$. The IMCR comparison of our framework with the two baseline methods, with the same parameter initialization, is shown in Fig.~\ref{fig:both_imcr} (b). The runtime of our framework, the uniform-weights, and the all-times baselines are $739s$, $727s$, and $9747s$, respectively. From Fig.~\ref{fig:both_imcr} (b), it is clear that the IMCR of our framework is better than the uniform-weights method, all over the horizon of signals. Although the all-times baseline obtains better IMCR than our approach, its runtime and memory consumption are noticeably larger, as it generates 61 STL formulas and learns the corresponding weight distributions.

\vspace{-2mm}
\section{Conclusion} \label{section:conclusion}
In this paper, we considered the problem of predicting the labels of prefix signals over time, given a dataset of labeled signals. Our proposed framework combines temporal logics and neural networks to construct a predictor for classifying the prefix signals. The effectiveness of our method was evaluated in an urban-driving and a naval surveillance scenario. In future work, we will explore advanced signal analysis techniques to find decision times. We will also explore other classification techniques as alternatives to the decision tree method, and evaluate their prediction performances.

\bibliography{references}

\end{document}